\documentclass[sigconf]{acmart}

\AtBeginDocument{%
  \providecommand\BibTeX{{%
    \normalfont B\kern-0.5em{\scshape i\kern-0.25em b}\kern-0.8em\TeX}}}

\setcopyright{rightsretained}
\copyrightyear{2022}
\acmYear{2022}
\acmDOI{}

\acmConference[SiKDD '22]{Ljubljana '22: Slovenian KDD Conference on Data Mining and Data Warehouses}{October, 2022}{Ljubljana, Slovenia}
\acmBooktitle{Ljubljana '22: Slovenian KDD Conference on Data Mining and Data Warehouses, October, 2022, Ljubljana, Slovenia}
\acmPrice{}
\acmISBN{}

\usepackage{listings}
\usepackage{xcolor}
\usepackage{adjustbox}
\usepackage{float}
\usepackage{color,soul}
\usepackage{aliascnt}
\usepackage{amsmath}
\usepackage{textgreek}
\usepackage{caption}
\usepackage{tikz-cd}
\usepackage{graphicx}
\usepackage{multirow}
\usepackage{hyperref}
\usepackage{academicons}
\restylefloat{table}

\begin{document}

\title{Forecasting Sensor Values in Waste-To-Fuel Plants: a Case Study.}

\author{Bor Brecelj}
\authornote{Both authors contributed equally to this research.}
\orcid{0000-0002-9670-3637}
\affiliation{%
  \institution{University of Ljubljana, Faculty of Mathematics and Physics}
  \streetaddress{Jadranska ulica 19}
  \city{Ljubljana}
  \country{Slovenia}
  \postcode{1000}
}
\email{bor.brecelj@gmail.com}

\author{Beno \v{S}ircelj}
\authornotemark[1]
\orcid{0000-0003-4994-067X}
\affiliation{%
  \institution{Jo\v{z}ef Stefan International Postgraduate School}
  \streetaddress{Jamova 39}
  \city{Ljubljana}
  \country{Slovenia}}
\email{beno.sircelj@ijs.si}

\author{Jo\v{z}e M. Ro\v{z}anec}
\orcid{0000-0002-3665-639X}
\affiliation{%
  \institution{Jo\v{z}ef Stefan International Postgraduate School}
  \streetaddress{Jamova 39}
  \city{Ljubljana}
  \country{Slovenia}}
\email{joze.rozanec@ijs.si}

\author{Bla\v{z} Fortuna}
\orcid{0000-0002-8585-9388}
\affiliation{%
  \institution{Qlector d.o.o.}
  \streetaddress{Rov\v{s}nikova 7}
  \city{Ljubljana}
  \country{Slovenia}}
\email{blaz.fortuna@qlector.com}

\author{Dunja Mladeni\'{c}}
\orcid{0000-0003-4480-082X}
\affiliation{%
  \institution{Jo\v{z}ef Stefan Institute}
  \streetaddress{Jamova 39}
  \city{Ljubljana}
  \country{Slovenia}}
\email{dunja.mladenic@ijs.si}

\renewcommand{\shortauthors}{Brecelj et al.}

\begin{abstract}
In this research, we develop machine learning models to predict future sensor readings of a waste-to-fuel plant, which would enable proactive control of the plant's operations. We developed models that predict sensor readings for 30 and 60 minutes into the future. The models were trained using historical data, and predictions were made based on sensor readings taken at a specific time. We compare three types of models: (a) a n\"aive prediction that considers only the last predicted value, (b) neural networks that make predictions based on past sensor data (we consider different time window sizes for making a prediction), and (c) a gradient boosted tree regressor created with a set of features that we developed. We developed and tested our models on a real-world use case at a waste-to-fuel plant in Canada. We found that approach (c) provided the best results, while approach (b) provided mixed results and was not able to outperform the n\"aive consistently.
\end{abstract}




\maketitle



\section{Introduction}
There is a wide range of applications of ML (machine learning). One of them is the modeling and control of chemical processes, such as the production of biodiesel. Introducing machine learning to such processes can improve quality and yield and help engineers predict anomalies to control the factory better. 

We modeled the JEMS waste-to-fuel plant, which produces high-quality diesel from organic waste. The plant has numerous sensors that measure temperature, and pressure, among other variables. It is operated by experts who must control the process. Since the chemical process is complex and, therefore, difficult to control, we built forecasting models that can predict future sensor readings based on historical data and the current state of the plant.

The model will be used to give plant operators additional information about the future state of the plant, which will allow them to make an informed decision about changing the plant's parameters and, therefore, adjust the process before it is too late.

\section{Related Work}
Organic wastes in energy conversion technologies are an active area of research aimed at reducing dependence on fossil fuels, optimizing production costs, improving waste management, and controlling emissions. Biochemical, physiochemical, and thermochemical processes produce different biofuels, such as bio-methanation, bio-hydrogen, biodiesel, ethanol, syngas, and coal-like fuels, which are studied by Stephen et al. ~\cite{biofuels1}. Work is also being done on optimization, such as catalyst selection, reactor design, pyrolysis temperature, and other important factors~\cite{biofuels2}.

Many ML methods have been developed to address waste management and proper processing for biofuel production, focusing on energy demand and supply prediction~\cite{biofuels3}. Aghbashlo et al.~\cite{biofuels4} provided a systematic review of various applications of ML technology with a focus on ANN (Artificial Neural Network) in biodiesel research. They provided an overview of the use of ML in modeling, optimization, monitoring, and process control. Models that predict the conditions of the biofuel production process that have the highest yield were created by Kusumo et al.~\cite{kusumo2017optimization} and Abdelbasset et al.~\cite{biofuels5}. The models used in these studies were kernel-based extreme learning machines, ANN, and various ensemble models.


\section{Use Case}

The JEMS waste-to-fuel plant produces synthetic diesel (SynDi) from any hydrocarbon-based waste, such as wood, biomass, paper, waste fuels and oils, plastics, textiles, rubber, and agricultural residues. The plant uses a chemical-catalytic de-polymerization process, the advantage of which is that the temperature is too low to produce carcinogenic gasses. It operates continuously and produces about 150 liters of fuel per hour. Although it uses the latest software available and allows remote control, there is no anomaly detection, prediction, or optimization. As a result, there is a great need for better understanding, optimization, and decision-making, given data availability. The company plans to sell and install over 1.500 SynDi systems over the next ten years. In practice, this means many SynDi plants in different locations worldwide. 

There are three main chambers in the pipeline, which are named B100, B200, and B300. The plant can be conceptually split into four stages
\begin{enumerate}
\item Feedstock inspecting and feeding;
\item Drying and mixing (chamber B100);
\item Processing (chamber B200);
\item Distilling (chamber B300).
\end{enumerate}
Since there are no sensors in the feedstock inspecting and feeding stage, we focused on the later stages, each of which takes place in one of the main chambers. 

In the drying and mixing stage (B100), the starting material is mixed with process oil, lime, and catalyst and is heated. During mixing, the material is broken down into smaller particles, and the water is evaporated. The primary chemical reaction occurs in the processing stage (B200). The material is fed to a turbine, and the reaction product evaporates through the diesel distillation column. If the diesel obtained is not of sufficient quality, it is redistilled in the second distillation stage (B300).


Currently, the plants are operated with highly skilled personnel and high costs for personnel training. Implementing automation, remote control, optimization, and interconnection among the plants would greatly facilitate their operation. Therefore, the main challenge to be solved by integrating AI is the self-control of the chemical process and the plant itself by minimizing the human resources required to operate the plants. Furthermore, operating many SynDi plants also means a significant challenge for ensuring remote control for troubleshooting, maintenance, and repair. AI integration aims to minimize the workforce required to operate the plants, minimize the resulting downtime due to human interaction, enable self-control and predictive maintenance of the SynDi plants, and achieve less downtime and higher production efficiency.

In modeling the waste-to-fuel processes, we decided to model each chamber separately. No model was developed for chamber B300 because it was not active during the period for which we obtained the data. As described above, a second distillation of the fuel is performed in chamber B300 only if the fuel in chamber B200 is not pure enough.

\section{Methodology}\label{S:METHODOLOGY}

\subsection{Data analysis}

The sensor measurements are from the experimental JEMS plant, which is located in Canada. The data consists of 154 sensors from January 2016 to January 2017. The measurements are taken at one-minute intervals and mostly measure temperature or pressure, but there are also sensors for motor current and valve position, among others. Since the data is from the prototype version of the waste-to-fuel plant, it contains many missing values. Our data set contained an average of 61.607 data points per sensor. We discarded all sensors with less than 6.000 data points and kept only those that corresponded to chambers B100 and B200, giving us data from 39 sensors.

Analysis of the dataset we received revealed that many values were missing. In particular, we noted that there were day-long intervals with a tiny number of measurements. We also noticed that specific sensor values remained constant at low temperatures~-~a condition best described by the waste-to-fuel plant's inactivity. We, therefore, decided to remove such values. Because there were many ten-minute gaps, we decided to resample the data at fifteen-minute intervals, taking the last value of each interval and assuming that conditions had not changed in the short time since the last measurement~-~a reasonable assumption for sensor values. The resulting data set contained an average of 7.884 data points per sensor.

We divided the dataset into a train and a test dataset, split on October 31\textsuperscript{st} 2016. The resulting train set included a total of 11.000 samples, and the test set included 3.000 samples.

\subsection{Model training}
In this research, we compare models that we develop using two different approaches. We first tried the neural network approach, in which the model makes predictions based only on sensor readings from the last five hours. Since the model did not perform better than the baseline, we began the second approach, developing features to describe the time series and capture its patterns. We used linear regression and gradient-boosted tree regressor. All the developed models were compared with the last-value model, which we used as a benchmark.

\subsubsection{Neural network approach}

We used the model developed for forecasting T\"upras' sensor values. T\"upras is an oil refinery, which is very similar to the JEMS use case. The model was used to forecast sensor values in different units of LPG production. Some of T\"upras' units are distillation columns, similar to JEMS' chamber B200. The model takes only past sensor values as input and predicts values for the future together with the prediction interval. More specifically, it predicts 10\textsuperscript{th}, 50\textsuperscript{th} and 90\textsuperscript{th} percentile, which is the case in all our models that give prediction interval.   

\begin{figure}[htb]
\includegraphics[width=0.35\linewidth]{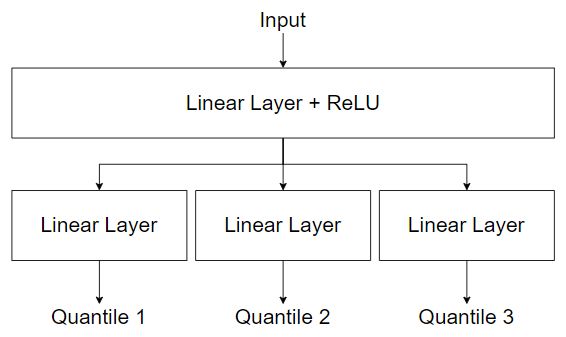}
\caption{Architecture of the neural network model, which gives the prediction interval.}
\label{fig:nn_architecture}
\end{figure}

Figure \ref{fig:nn_architecture} shows the architecture of the neural network. The model is a feedforward neural network with two layers. First, there is a linear layer with ReLU activation. The second layer has a separate linear layer for each quantile. The hidden dimension of the model is calculated from the number of features and the number of targets using the formula 
$\lfloor \frac{n_\mathrm{features}}{2} \rfloor + n_\mathrm{targets}$.

During training, we used the quantile loss function, which is defined as
$$
\max \left\{ q \cdot \left( y_\mathrm{true} - y_\mathrm{pred} \right),
\left(1-q\right) \cdot \left( y_\mathrm{pred} - y_\mathrm{true} \right) \right\},
$$
where $q$ is the observed quantile (in our case, it can be $0.1$, $0.5$ or $0.9$), $y_\mathrm{true}$ is the true target value and $y_\mathrm{pred}$ is the corresponding quantile of the prediction. In the case of $q = 0.5$, the loss is equal to the mean absolute error divided by two. When calculating the loss of 10\textsuperscript{th} percentile ($q=0.1$), a prediction that is greater than the true value is heavily penalized, while a prediction that is lower than the true value has a smaller loss and is therefore encouraged.

The model is implemented in the PyTorch library~\cite{pytorch}. Since sensors measure different quantities, the values have to be scaled before learning. Here we used Min-Max scaler from the scikit-learn library, scaling all values between zero and one.

\subsubsection{Feature engineering}

The neural network model described above did not outperform the benchmark model. As a result, we decided to try another approach, where we developed features that better describe past sensor values and capture their patterns. One of the problems of the neural network model was that it had too many features. We decided to build a separate model for each sensor to tackle this problem. Each model uses only features calculated from the values of the sensor being predicted.

With the help of plant operators, we decided to consider at most five hours of data before the prediction point to issue a forecast. Since the latest data is usually more important in determining future sensor values, we created features on seven different time windows: 30, 45, 75, 120, 180, 240, and 300 minutes. For each time window, we computed the following features:
\begin{itemize}
    \item average sensor value,
    \item fraction of peaks in the window,
    \item percentage change between first and last value in the time window,
    \item slope (coefficient of the least squares line through the points in the window),
    \item simple prediction (extension of the least squares line to the future),
    \item slope ratio (slope on the smaller window divided by the slope on the bigger window).
\end{itemize}
Besides features mentioned above, which depend on the window size, we also included features that were calculated only on the biggest time window (300 minutes):
\begin{itemize}
    \item last value,
    \item maximal value,
    \item last value relative to the maximal value.
\end{itemize}
The features above attempt to capture different time series characteristics: 
\begin{itemize}
    \item \textit{trend}: described by percentage change and slope;
    \item \textit{growth pattern}: described by the fraction of peaks, which indicate whether the growth is steady or it has ups-and-downs. Furthermore, the slope indicates how aggressive such growth is;
    \item \textit{expected value}: an approximation of the expected value is given through the average, last value, maximal value, and simple prediction.
\end{itemize}

Using developed features, we trained a linear regression model, and a gradient boosted tree regressor from the CatBoost library~\cite{catboost}. We used root mean squared error (RMSE) for the loss function.

\section{Results and Analysis}

We built models for main chambers B100 and B200 with two forecasting horizons (30 and 60 minutes). Tables \ref{table:resultsB100} and \ref{table:resultsB200} show mean squared error (MSE) and mean absolute error (MAE) on chambers B100 and B200, respectively. There are three different neural network models (NN), which differ in the size of the window from which it gets the data.

\begin{table}[htb]
\centering
\begin{tabular}{ c || c | c || c | c }
  & \multicolumn{2}{c||}{horizon = 30min} &
  \multicolumn{2}{c}{horizon = 60min} \\
  & MSE & MAE & MSE & MAE \\ 
 \hline
 last-value model & 21.0533 & \textbf{1.4320} & 50.6636 & 2.5128 \\
 NN, window = 5h & 21.7525 & 1.6512 & 47.0545 & 2.5413 \\
 NN, window = 3h & 19.7441 & 1.6109 & 45.3450 & 2.4127 \\
 NN, window = 2h & 18.9717 & 1.6023 & 46.5047 & 2.5357 \\
 Linear regression & 19.4264 & 1.4634 & 49.2268 & 2.5145 \\
 Catboost & \textbf{16.9030} & 1.4478 & \textbf{38.3066} & \textbf{2.3164} \\
 \end{tabular}
\caption{MSE and MAE on the test set of models when predicting for chamber B100.}
\label{table:resultsB100}
\end{table}

\begin{table}[htb]

\begin{tabular}{ c || c | c || c | c }
  & \multicolumn{2}{c||}{horizon = 30min} &
  \multicolumn{2}{c}{horizon = 60min} \\
  & MSE & MAE & MSE & MAE \\ 
 \hline
 last-value model & 52.3380 & \textbf{2.0577} & 124.9735 & \textbf{3.3768} \\
 NN, window = 5h & 69.4678 & 3.8227  & 129.0330 & 4.9927 \\
 NN, window = 3h & 57.9902 & 3.3601 & 121.1315 & 4.7431  \\
 NN, window = 2h & 55.8769 & 3.1797 & 117.4154 & 4.7146  \\
 Linear regression & 55.0218 & 3.2293 & 115.7457 & 4.5888 \\
 Catboost & \textbf{49.3329} & 2.5305 & \textbf{109.5303} & 3.9745  \\

 \end{tabular}
\caption{MSE and MAE on the test set of models when predicting for chamber B200.}
\label{table:resultsB200}
\end{table}

From the tables \ref{table:resultsB100} and \ref{table:resultsB200} we can see that the five-hour window's neural network performed worse than the benchmark. The main reason for such poor results was too many features for the amount of data that we have. More precisely, the neural network model uses the values of all sensors in the chamber we are predicting. This means that there are six hundred features resulting in more than two hundred thousand trainable parameters for the model of chamber B200. We also have to consider that the neural network predicts future sensor values and prediction intervals. Therefore, there are too many features and target values for the amount of data that we have.

We included results of two more neural network models with three hours and two-hour windows since reduced window size results in a smaller number of features and trainable parameters. For example, the neural network model with a two-hour time window for chamber B200 had two hundred and forty features and almost fifty thousand trainable parameters. Neural network models with smaller window sizes performed better, which confirms that we had too many features. 

The features that we developed using the second approach were used with two models, linear regression and the Catboost model. Comparing those two models, the Catboost model performed better because it can capture more than just linear relationships between the features and the target. The Catboost model also outperformed the neural networks, where one of the main differences is that the neural network uses all sensors from the chamber while the Catboost model uses only sensor values of the sensor which is being predicted. This results in forty-five features for the model that predicts one sensor, which solves the problem of too many features. In addition, the Catboost model produced better results than the benchmark when comparing the mean squared error (MSE). During the training, we used RMSE as a loss function, meaning that RMSE was minimized and, therefore, also MSE.  

The tables show that although most models outperform the benchmark regarding MSE, almost all of them do not surpass the benchmark when considering MAE. When measuring MSE, predictions with strong spikes where such spikes do not take place are penalized more. Therefore, models with a competitive MSE are considered to rarely predict spikes when such spikes do not take place. This is a key feature for our use case, given that we are interested to understand whether an irregularity will take place or not. Therefore, the models give valuable information even though the average prediction is not entirely accurate.

\begin{figure}[htb]
\includegraphics[width=0.450\linewidth]{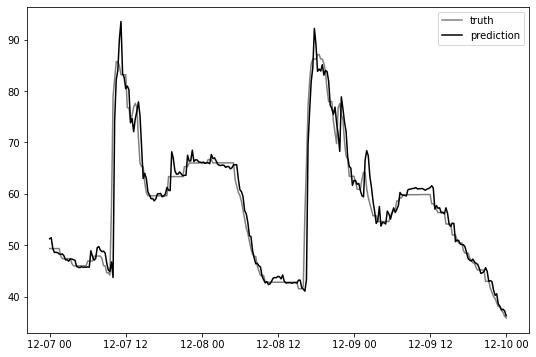}
\caption{True value and prediction of the Catboost model for a temperature sensor in chamber B100.}
\label{fig:prediction_catboost_b100}
\end{figure}

\begin{figure}[htb]
\includegraphics[width=0.450\linewidth]{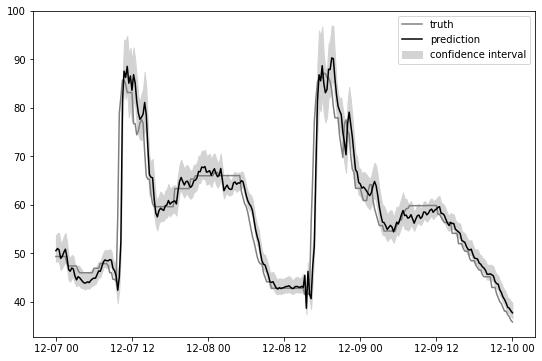}
\caption{True value and prediction with a confidence interval of the neural network model with a two-hour window for a temperature sensor in chamber B100.}
\label{fig:prediction_nn_b100}
\end{figure}

Figure \ref{fig:prediction_catboost_b100} shows the Catboost model prediction on the test set together with the true values of the temperature sensor in chamber B100. The neural network model's prediction of the same sensor is presented in Figure \ref{fig:prediction_nn_b100}. Since the neural network model also outputs prediction interval, it is shown in the abovementioned Figure.

From the plots, we can see that both models can closely predict future sensor values. In the case of the neural network model, the actual value is mainly inside the predicted confidence interval, except when there is a significant change in the sensor value. However, there is no problem with models not being able to predict significant changes resulting from a manual change in plant setpoint parameters, which our data does not capture. Overall, we consider the best model was the Catboost model, given in all cases it outperformed the rest of the models when considering MSE, and also achieved the best MAE when predicting chamber B100 with a time horizon of 60 minutes.

\section{Conclusion}
We compared a set of models to predict sensor values for a waste-to-fuel plant: a neural network, linear regression, gradient-boosted tree regressor, and the last-value model. The last-value model was used as a benchmark. We developed three neural network models which were different in time window size. The neural network models were built based on the hypothesis that a simple neural network and raw sensor readings as features are enough to model the process. The results showed that this is not the case because the process is too complicated for the amount of data that we obtained. Lastly, we used feature engineering to develop features that better describe the time series. Features were used for learning linear regression, and the gradient boosted tree regressor, where the latter produced the best results in our case.

\begin{acks}
This work was supported by the Slovenian Research Agency and the European Union’s Horizon 2020 program project FACTLOG under grant agreement number H2020-869951.
\end{acks}

\bibliographystyle{ACM-Reference-Format}
\bibliography{main}

\end{document}